\documentclass[]{elsarticle}

\usepackage{hyperref}

\journal{}

\usepackage{amsmath}
\usepackage{amsfonts}
\usepackage[ruled, lined, boxed, linesnumbered]{algorithm2e} 
\usepackage{color}
\usepackage{float}
\usepackage{graphicx}
\usepackage{hyperref}
\usepackage{caption}
\usepackage{subcaption}
\usepackage[utf8]{inputenc}
\usepackage[colorinlistoftodos]{todonotes}

\usepackage{fancyhdr}
\newcommand{\bphi}{\boldsymbol{\phi}}
\newcommand{\btheta}{\boldsymbol{\theta}}
\newcommand{\bgamma}{\boldsymbol{\gamma}}
\newcommand{\htheta}{{\boldsymbol{\theta}}}
\newcommand{\halpha}{{\boldsymbol{\alpha}}}
\newcommand{\bx}{\mathbf{x}}
\newcommand{\bxr}{\mathbf{x}^{\textrm{ref}}}
\newcommand{\bxa}{\mathbf{x}^{\textrm{alt}}}

\newcommand{\define}{\overset{\Delta}{=}}

\newtheorem{definition}{Definition}[section]
\newtheorem{remark}{Remark}[section]

\SetKwFunction{Sample}{Sample}
\SetKwFunction{Entropy}{Entropy}
\SetKwFunction{MI}{MI}
\SetKwFunction{MC}{MC}
\SetKwFunction{Trial}{Trial}
\SetKwFunction{Design}{TrialDesign}

\begin{document}

\begin{frontmatter}

\title{On Preference Learning Based on Sequential Bayesian Optimization with Pairwise Comparison}


\author[mymainaddress]{Tanya~Ignatenko\corref{mycorrespondingauthor}}
\ead{tignatenko@gnhearing.com}

\author[mymainaddress]{Kirill~Kondrashov}
\ead{kkondrashov@gnhearing.com}

\author[mymainaddress]{Marco Cox}
\ead{extmcox@gnhearing.com}

\author[mymainaddress]{Bert de Vries}
\ead{bdevries@gnresound.com}

\address[mymainaddress]{R\&D Global Research, GN Group, John F Kennedylaan 2, 5612 AB Eindhoven}



\cortext[mycorrespondingauthor]{Corresponding author}


\begin{abstract}
User preference learning is generally a hard problem. Individual preferences are typically unknown even to users themselves, while the space of choices is infinite. Here we study user preference learning from information-theoretic perspective. We model preference learning as a system with two interacting sub-systems, one representing a user with his/her preferences and another one representing an agent that has to learn these preferences. The user with his/her behaviour is modeled by a parametric preference function. To efficiently learn the preferences and reduce search space quickly, we propose the agent that interacts with the user to collect the most informative data for learning. The agent presents two proposals to the user for evaluation, and the user rates them based on his/her preference function. We show that the optimum agent strategy for data collection and preference learning is a result of maximin optimization of the normalized weighted Kullback-Leibler (KL) divergence between true and agent-assigned predictive user response distributions. The resulting value of KL-divergence, which we also call remaining system uncertainty (RSU), provides an efficient performance metric in the absence of the ground truth.   This metric  characterises how well the agent can predict user and, thus, the quality of the underlying learned user (preference) model. Our proposed agent comprises sequential mechanisms for user model inference and proposal generation. To infer the user model (preference function), Bayesian approximate inference is used in the agent. The data collection strategy is to generate proposals, responses to which help resolving  uncertainty associated with prediction of the user responses the most. The efficiency of our approach is validated by numerical simulations. Also a real-life example of preference learning application is provided.

\end{abstract}


\end{frontmatter}


\thispagestyle{fancy}

\section{Introduction}
In this work, we study the problem of preference learning. We start with a motivating example presenting a preference learning problem in the context of the fitting a hearing aid (HA). 
This example allows us to set the scene for the preference learning problem and reason about the properties and requirements of the corresponding solution.   

\textbf{Motivating Example:}
Fitting and tuning of HAs has always been considered a tedious task of healthcare professionals (HCPs). Traditional approaches for fitting HA parameters rely on compensation of a user’s hearing loss, based on audiograms, by applying rules such as NAL-NL1 or NAL-NL2 \cite{Keidser2011}. These rules, however, do not take into account specific user preferences. Tuning of the HA parameters would happen at some later moment of time, at an HCP's office, when a user shares some qualitative feedback about his/her experience of using their HAs with the HCP. The tuning result heavily depends on the HCP's experience and expertise levels, and the degree to which the user feedback can be related to the required tuning. 
In practice, due to generic fitting and tuning procedures, HA users remain not fully satisfied with the result that leads to recurrent visits to HCPs or less frequent HA use. This gives a rise to optimal personalization problem of HA parameter settings. It is desirable to empower the  user  to  take  direct  decisions  and  have  direct  impact on the  tuning  process by creating an automated system for learning user preferences in an efficient and minimally obtrusive way.

In the HA example, equalizer tuning, simulations, rendering and other similar applications, the result of parameter tuning cannot be objectively evaluated and depends on an individual perceptual subjective evaluation. Individual preferences are often unknown even to the users themselves and might require infinite number of trying different options to discover them.  Moreover, from the user perspective, the relation between the tuning process and its outcome is not straightforward. Indeed, there are typically many interconnected parameters that have a nonlinear relationship with the final result. Therefore, in this work, we study a problem of efficient automated search for optimal parameters or, more generally, user preference by incorporating a user feedback into the learning cycle. From psychological prospective, see, e.g., \cite{kendall1990correlation}, people find it easier to compare two options and indicate their preference for one of them,  rather than to rate options according to a numeric scale. Therefore, to lower the burden on the user and guarantee reliability of the feedback, we consider the setting, where the  user is required  to  evaluate  a sequence of pairs of proposals, by stating his/her\footnote{Later in the text, for compactness, we use ``his'' as a gender-neutral pronoun associated with a ``user'' .}  choice  for  one  of  them.

The problem of preference learning has been an active research topic in the past years. Brochu et. al \cite{Brochu2007} studied active preference learning with Gaussian processes used to model preference function and expected improvement criterion as acquisition function for efficient data collection. A good overview of preference learning using Gaussian processes and further references are presented in \cite{brochu2010tutorial}.  Gonz\'{a}lez et. al \cite{gonzalez2017a} studied the problem of preferential Bayesian optimization of a black-box function with duelling bandits. In their work an unknown function is also modeled using Gaussian processes and considered acquisition strategies included expected improvement and duelling-Thompson sampling. Petrus et. al \cite{Petrus2020} studied the preference learning problem by incorporating expert knowledge, where the feedback provided to the agent is  optimal projection information provided by the experts, and thus is purely instructive.  Another direction of preference function modeling was taken in \cite{cox_parametric_2017}, where parametric user preference models were proposed. In their work, Thompson sampling together with Bayesian approximate inference was used to learn the user preferences and cumulative value of the preference function was used to assess the learning performance. It was shown that parametric approach could outperform preference function models based on Gaussian processes. 
Remarkably, in situations with preference learning, the optimal user preferences as well as the corresponding value of the preference function are unknown. Therefore, in the approaches based on the expected improvements, the current estimate of the user preference  function for the optimal preference is used as a ground truth, which need not be close to the actual user preference. Hence, the procedure does not guarantee convergence to the true preference. Duelling approaches only take into account the second order statistics for defining acquisition strategies. Moreover, performance metrics such as expected improvement, valuation function or Copeland score do not guarantee convergence to the optimum. 

In Bayesian optimization literature,  other alternatives for acquisition functions were proposed that included entropy search (ES) \cite{Hennig2012EntropySF} and predictive entropy search (PES) \cite{Lobatob2014} approaches. ES-based acquisition function, however, relies on the entropy reduction relative to the uniform distribution and thus does not guarantee efficient exploration of the search space and correct localization of the user preferences. PES approach aims at exploring the data points that provide the highest information gain about location of the global maximum of some unknown function. Similar to ES and PES approaches, we study the problem of the preference learning from the information-theoretic perspective. 
While our approach results into the acquisition function similar in form to the PES one, it is conceptually very different. We provide theoretical foundation for the optimal agent strategy that combines both preference learning and data acquisition, and propose corresponding performance measure. Instead of using estimates of unobservable preference function values, we concentrate on the information provided in the user responses, in order to steer the learning process. The proposed performance measure can be efficiently used to evaluate the learning performance in the absence of true user preference information. 

\subsection{Summary of Our Contribution}
\begin{figure}[!tb]
   \centering
   \includegraphics[scale=0.5]{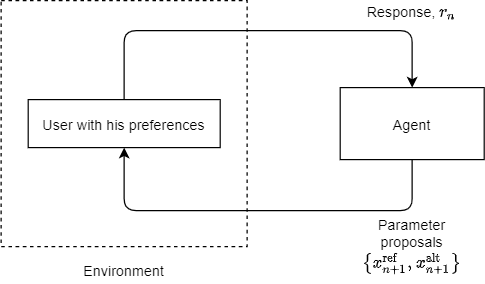}
    \caption{A preference learning system, as two interacting sub-systems, i.e., a user and an agent. For each interaction, the agent generates two proposals. The user evaluates these two proposals, based on his preferences, and provides the  ranking of the presented proposals (binary response). }
    \label{fig:ha_sys}
\end{figure}

The goal of this work is to design a Bayesian optimizer for user preference learning from binary choices. This optimizer has to find the optimal preferences of the user, given binary user responses, in as few trials as possible. Binary user responses correspond to user indication of his preference for one of the two presented choices.  Our approach is based on a parametric user preference model, proposed in \cite{cox_parametric_2017}. In this case, the optimal preference can be expressed as the location of the maximum of the latent user preference function that we need to infer. 
We take an information-theoretic point of view to the problem. We model the preference learning system as two interacting sub-systems, where one sub-system represents the user with his preferences and the other represents an agent that probes the user, such that it can efficiently learn the user preferences, see Fig.\ref{fig:ha_sys}.  The agent strategy is based on sequential reduction of uncertainty about the user after each interaction between the user and the agent. We show that the optimal strategy is the result of maximin optimization of the normalized weighted Kullback-Leibler (KL) divergence between true and agent-assigned predictive user response distributions. This approach results into a theoretically-grounded acquisition function. Moreover, we use the value of the optimized KL-divergence as a metric that characterizes the quality of the learned preference and allows us to monitor the learning process and its convergence. This metric does not require the knowledge of the true preference. The optimal agent strategy and the metric are derived based on the concepts of universal prediction, see e.g. \cite{Merhav98universalprediction}. Finally, our learning procedure is sequential, sometimes in the literature referred to as online. The advantage of this approach is two-fold: firstly, we do not need to store all data required for the model update, and, secondly, this approach allows capturing the dynamics of (possibly) changing user preference.

\textit{Notations:}
In the rest of the text, we denote random variables  by capital letters, $R,$ realizations or observations by small letters, $r,$ and vectors by bold letters, $\boldsymbol{r}$. Moreover, sequences are denoted by $r_1^n$  and $\{a,b\}_1^{n}$, and defined as $r_1^n=\{r_1,r_2,\ldots,r_n\}$ and 
$\{a,b\}_1^{n}=\{\{a_1,b_1\},\{a_2,b_2\},\ldots,$ $\{a_{n},b_{n}\}\},$ respectively.

\section{User and Preference Function Modeling}
We start with specifying the user model, which represents a probabilistic description of user responses, when comparing two proposals. The proposals are expressed as sets of parameters defining an outcome that user evaluates. E.g., in the motivating example, parameters are represented by set of HA parameters that are used by a HA device to process sound, which is evaluated by a user. Thus, parameters are latent random variables in the models we propose. Throughout the text, we use parameter proposals, parameters and proposals interchangeably, and it will be clear form the context if proposals refer to the parameters or outcome of parameter application.  Integral parts of the model include a user preference function and a distribution of the user responses  to the presented choices of parameters.

Consider a vector of parameters, defined on a $D$-dimensional continuous compact surface.
In particular, parameters $\bx$ are defined on a $D$-dimensional hyper-cube, i.e., $\bx \in [0, 1]^D$. We can always guarantee this by normalizing the actual parameters by their physical range. Our goal is to find optimal values of parameters, denoted by $\htheta,$ corresponding to the preference for a particular user. Here we follow the model setting of ~\cite{cox_parametric_2017}.

The user model is based on two assumptions: (1) a user preference exists and is described by a unimodal preference function $f(\cdot)$; and (2) a user can be uncertain about his preference, which can be expressed in terms of an additive random variable. 

In general, the user preference function is unknown. We assume this latent  preference function $f(\bx;\htheta,\Lambda)$ to be a parametric function of parameters, $\mathbf{x}\in [0,1]^D$ with known form but unknown shape. This shape is characterized by  \textit{tuning parameters},  $\htheta\in [0,1]^D$ and  $\Lambda=\text{diag}([\lambda_1,\lambda_2,\ldots,\lambda_D]), \lambda_d\in \mathbb{R}^{+}, d=1,2,\ldots , D$. In particular, the user preference function is specified by the following function  
\begin{equation}\label{eq:pref_function}
f(\mathbf{x};\btheta,\Lambda) = -\sqrt{(\mathbf{x}-\btheta)^T \Lambda (\mathbf{x}-\btheta)}.
\end{equation}
Note that the tuning parameters correspond to 
the location of the function optimum, $\htheta,$ that is, in fact, the optimal parameters corresponding to the user preference; and to the spread around this optimum, $\Lambda,$ that characterizes user sensitivity to parameter changes, respectively. The tuning parameters  are user-specific and need to be learned. 
The form of this preference function has an attractive property that it allows for quick narrowing of the search space, see ~\cite{cox_parametric_2017}. 

Next, in order to model user preference uncertainty, we define a value function $u(\cdot)$ as
\begin{equation}\label{eq:val_fucntion}
u(\mathbf{x})
=f(\mathbf{x};\htheta,\Lambda)+\varepsilon,
\end{equation}
where the user uncertainty error $\varepsilon$ is assumed to have the standard Gaussian distribution, i.e.,
\begin{equation}\label{eq:user_uncert}
\varepsilon \sim \mathcal{N}(0,1).
\end{equation}
The choice for this assumption is further discussed after we introduce the notion of the user response.

Consider now a user response to a pairwise comparison of parameters from a trial, defined by a pair $\{\bxr,\bxa\}$, where $\bxr$ and $\bxa$ are the so-called \textit{reference and alternative  parameter proposal}, respectively. We call reference proposal $\bxr$  to be the highest rated proposal from the previous trial and this situation is challenged by the alternative proposal $\bxa$, in the given trial.
The user response depends on his preferences for the presented parameters, which are given by the value function for these parameters. The user response, $R,$ takes values from $\{0,1\}$, where response $1$ to a trial $\{\bxr,\bxa\}$ indicates that parameter $\bxa$ is preferred over or is as good as $\bxr,$ and $0$, otherwise, i.e.,
\begin{equation}\label{eq:response}
R\define \left\{\begin{matrix}
1,& \textrm{if } \bxa \succeq \bxr\\
0,& \textrm{if } \bxr \succ \bxa
\end{matrix}\right.,
\end{equation}
here $\succeq$ denotes preferred or equivalent to, and $\succ$ denotes strict preference. 

Now, the probability of a positive user response, i.e., that $R$ takes on value $1$, 
in a trial $\{\bxr,\bxa\}$, is  given by
\begin{eqnarray}\nonumber
&&\hspace{-50pt}\Pr\{R=1\,|\,\{\bxr,\bxa\}, f\}\\\nonumber 
&=&\Pr\{u(\bxa) \geq u(\bxr)\}\\\nonumber
&=& \Pr\{f(\bxa;\htheta,\Lambda)+\varepsilon'
 \geq f(\bxr;\htheta,\Lambda)+\varepsilon\}\\\nonumber
 &=& \Pr\{\varepsilon-\varepsilon'
 \leq f(\bxa;\htheta,\Lambda)- f(\bxr;\htheta,\Lambda)\} \\\nonumber
 &=&\Phi\left(\frac{ f(\bxa;\htheta,\Lambda) - f(\bxr;\htheta,\Lambda)}{\sqrt{2}}\right)\\
&=&\Phi\left( f(\bxa;\htheta,\Lambda') - f(\bxr;\htheta,\Lambda')\right),\label{eq:response_par}
\end{eqnarray}
where  $\Phi(\cdot)$ is the cumulative distribution function (CDF) of the standard Gaussian distribution, and $\Lambda'=\frac{\Lambda}{2}$.  Thus, $R$ is a Bernoulli random variable with parameter $\Phi\left(f(\bxa;\htheta,\Lambda') - f(\bxr;\htheta,\Lambda')\right).$  Therefore, given the user preference and response model, defined by (\ref{eq:pref_function})-(\ref{eq:response_par}), the probability mass function (PMF) of a user response to a trial $\{\bxr,\bxa\}$ is given by
\begin{eqnarray}\nonumber
&&\hspace{-35pt}P\left(r\,|\,\{\bxr,\bxa\},f \right)\\\nonumber &=&\Phi\left(f(\bxa;\htheta,\Lambda') - f(\bxr;\htheta,\Lambda')\right)^r \\\nonumber
&&\hspace{15pt}\cdot \left(1-\Phi\left(f(\bxa;\htheta,\Lambda') - f(\bxr;\htheta,\Lambda')\right)\right)^{1-r}\\\nonumber
&=& \Phi\left(f(\bxa;\htheta,\Lambda') - f(\bxr;\htheta,\Lambda')\right)^r\\
&&\hspace{15pt}\cdot \Phi\left( f(\bxr;\htheta,\Lambda') - f(\bxa;\htheta,\Lambda')\right)^{1-r}.\label{eq:user_response}
\end{eqnarray}
This user response model, which  relates binary observations to the continuous latent function, is also known as  the Thurstone-Mosteller law of comparative judgment \cite{ashby_measurement_1959}. In statistics, it is called the binomial-probit regression model.  

\begin{remark} Note that we can assume unit-variance of the user uncertainty error because other values can be  absorbed by the preference function as a scaling factor in the user sensitivity. Indeed, when $\varepsilon \sim \mathcal{N}(0,\beta^{-1})$ 
the user response model becomes
\begin{eqnarray*}
&&\hspace{-35pt}P\left(r\,|\,\{\bxr,\bxa\},f \right) \\
&=&  \Phi\left( \frac{\left(f(\bxa;\htheta,\Lambda) - f(\bxr;\htheta,\Lambda)\right)}{\sqrt{2}\beta^{-1}}\right)^r\\
&&\hspace{10pt}\cdot \Phi\left( \frac{\left(f(\bxr;\htheta,\Lambda) - f(\bxa;\htheta,\Lambda)\right)}{\sqrt{2}\beta^{-1}}\right)^{1-r}\\
&=& \Phi\left(f(\bxa;\htheta,\Lambda'') - f(\bxr;\htheta,\Lambda'')\right)^r\\ 
&&\hspace{10pt}\cdot \Phi\left( f(\bxr;\htheta,\Lambda'') - f(\bxa;\htheta,\Lambda'')\right)^{1-r},
\end{eqnarray*}
where $\Lambda''=\frac{\beta^2\cdot\Lambda}{2}$. 
\end{remark}

\begin{remark} Observe that, in this model, we can re-define $\Lambda$ as  $\Lambda'$ or $\Lambda''$. Since $\Lambda$ is an unknown parameter, this scaling can be applied without loss of generality. However, care need to be taken while setting the priors for this parameter, taking into account physical interpretation of parameters in a specific application and thus possibly also re-scaling hyper parameters of the prior distributions.
\end{remark}

The following definition summarizes our user model.
\begin{definition}[User Model]
\label{def:usermodel}
A user model is defined by the PMF of the user response $r$ to a trial $\{\bxr,\bxa\}$ and is given by
\begin{eqnarray}\nonumber
&&\hspace{-24pt}P\left(r\,|\,\bxr,\bxa,f \right)
\,=\, \Phi\left(f(\bxa;\htheta,\Lambda') - f(\bxr;\htheta,\Lambda')\right)^r\\
&&\hspace{40pt} \cdot \Phi\left( f(\bxr;\htheta,\Lambda') - f(\bxa;\htheta,\Lambda')\right)^{1-r}\hspace{-15pt},\label{def:user_response}
\end{eqnarray}
where $\Phi(\cdot)$ is the CDF of the standard normal distribution. Here $f(\cdot;\cdot)$ is a user preference function, given by 
\begin{equation}\label{def:pref_function}
f(\mathbf{x};\btheta,\Lambda) = -\sqrt{(\mathbf{x}-\btheta)^T \Lambda (\mathbf{x}-\btheta)},
\end{equation}
the user response relation to a trial $\{\bxr,\bxa\}$ is defined by
\begin{equation}\label{def:response}
R\define \left\{\begin{matrix}
1,& \textrm{if } \bxa \succeq \bxr\\
0,& \textrm{if } \bxr \succ \bxa
\end{matrix}\right.,
\end{equation}
and the user uncertainty about his response to a given trial is characterized by the standard Gaussian distribution: 
\begin{equation}\label{def:user_uncert}
\varepsilon \sim \mathcal{N}(0,1).
\end{equation}
\end{definition}

Observe that the user model, defined as above, is a parametric model of the user preferences and sensitivity. Our goal is to estimate the user preference parameter $\htheta,$ based on a sequence of trials and corresponding user responses. This problem can be seen as a problem of finding the optimum of the latent user preference function, and thus requires learning the function or, equivalently, its parameters. Note, however,  that sensitivity parameter $\Lambda$ is, in fact, a nuisance parameter, since our primary goal is learning the location of the optimum.
Moreover, since we need to learn user preferences in as few trials as possible, this learning problem also relates to the problem of active learning. There, the goal is to learn a model efficiently by selecting the most relevant data (trials in our setting). 
Given unimodality of the preference function, it is also desirable to construct trials such that in each given trial, an alternative proposal is  better than a reference proposal with high probability. The  resulting procedure is then similar in flavor to gradient ascent.   

In the next section, we define the inference strategy for user preference learning, but, first, we  specify the full model for our preference learning system and its properties. 

\section{Preference Learning System and Inference Strategy}
\subsection{Parameter Transformation}
Before going into the model specification, we introduce useful parameter transformation and define the corresponding parameter vector:
\begin{eqnarray}\label{eq:phi-def}
\boldsymbol{\phi} &= &
\begin{bmatrix}\
\alpha_1\\
\vdots\\
\alpha_D\\
\gamma_1\\
\vdots\\
\gamma_D\\
\end{bmatrix},
\end{eqnarray}
where 
\begin{eqnarray}\label{eq:theta-to-alpha}
\alpha_d &=& \Phi^{-1}(\theta_d),\\
\gamma_d&=& \ln(\lambda_d), \hspace{14pt} d=1,2,\ldots, D,
\end{eqnarray}
and $\Phi^{-1}(\cdot)$ is the inverse of the CDF of the standard Gaussian distribution.  The latter non-linear monotonic transformation guarantees that the values of parameters $\htheta$ are constrained to the $[0,1]^D$ hyper-cube during the learning process, while we operate in the transformed continuous parameter space. From now on, we also assume that trial proposals are in the same transformed domain as the parameters.
Similarly, transformation of the user sensitivity  helps to constrain this parameter to the positive real subspace. Further, we re-define the preference function with this transformed parameter vector as
\begin{equation}\label{eq:parame_trans}
f(\bx;\bphi) \define f(\Phi(\bx);\Phi(\halpha),\exp(\Gamma)),    
\end{equation}
where the transformations are applied component-wise to $\halpha$ and $\bx$, thus, e.g., $\Phi(\halpha)$ denotes a vector  $\Phi(\halpha)=\{\Phi(\alpha_1),\Phi(\alpha_2),\ldots, \Phi(\alpha_D)\},$ and component-wise to the diagonal of $\Lambda,$ then $\exp(\Gamma)$ is a diagonal matrix with elements  $\exp(\gamma_d),$ $ d=1,2,\ldots, D$ on its diagonal.

\subsection{Preference Learning System Definition}
Our goal is to design an efficient system for learning parameters of the latent user preference function. We have two integral components here: a user with his preferences and an agent or optimizer that interacts with the user by offering parameter settings. Therefore, we model the full preference learning system as a combination of two sub-systems, representing a user and an optimizer, see Fig.~\ref{fig:ha_sys}.  In this system, the user can be seen as an environment in which the optimizer operates. Different settings for the user preference function, $\bphi$, correspond to different users or environments, and thus different user models.   User responses to the trials, $r,$  are observations of the environment. Moreover, they are reactions to the proposal trials.  An optimizer or agent is a part of the preference learning system that interacts with the environment by generating  trials or actions to which the environment reacts. The agent generates trials, according to some objective. The objective of our agent is to generate trials that result in accurate estimate of the user preferences for the parameters in as little number of trials as possible. 

\begin{definition}\label{def:HAsys}
A preference learning system is defined by a combination of two interacting sub-systems $\{P^a,P^u\}$, an agent and a user, given by a sequence of $n=1,2,\ldots$  distributions
\begin{eqnarray}\label{eq:agent_dist}
\hspace{-18pt}P_n^a&=& p( \{\bxr_n,\bxa_n\}| \bphi, r_1^{n-1}, \{\bxr,\bxa\}_1^{n-1})\\
\hspace{-18pt}P_n^u&=&P(r_n | \bphi, r_1^{n-1}, \{\bxr,\bxa\}_1^{n-1}, \{\bxr_n,\bxa_n\}),\label{eq:response_pred}
\end{eqnarray}
where $p(\{\bxr_n,\bxa_n\}|\, \bphi, r_1^{n-1}, \{\bxr,\bxa\}_1^{n-1})$ is
a generative PDF for trials $\{\bxr_n,$ $\bxa_n\}$ in  the environment $\bphi$, given the previous sequence of observations, $r_1^{n-1},$ and trials,    $\{\bxr,\bxa\}_1^{n-1}$; while $P(r_n | \bphi, r_1^{n-1}, \{\bxr,\bxa\}_1^{n-1}, \{\bxr_n,\bxa_n\})$ is a generative probability of the response in this environment to the new trial\\ $\{\bxr_n,\bxa_n\},$ given  $r_1^{n-1}$ and $\{\bxr,\bxa\}_1^{n-1}.$ 
\end{definition}

Observe that, for the agent, described by (\ref{eq:agent_dist}), observations $r_1^{n-1}$  are input data and actions $\{\bxr_n,\bxa_n\}$ are outputs, while for the user, described by (\ref{eq:response_pred}), the situation is reversed. 

In the preference learning system, defined as above, trials or actions are data that are produced by the agent, and every time a trial is presented, the user reacts to this \textit{given trial}. As such the relation between trials and responses to trials becomes causal. Therefore, interactions between the user and the agent are characterized by a sequence of distributions
\begin{eqnarray*}
p(\phi), p(\{\bxr_1,\bxa_1\}| \bphi), P(r_1 | \bphi,\{\bxr_1,\bxa_1\}),
p(\{\bxr_2,\bxa_2\}| \bphi, r_1, \{\bxr_1,\bxa_1\} ),\ldots
\end{eqnarray*}

\subsection{Agent Strategy Design}

Given the preference learning system, we have to deal with  two connected problems. One problem relates to inference of the parameters of the user preference function. We look closely at this problem in Section \ref{sec:UM_inference}. Another problem deals with  efficient learning, and thus experimental design that leads to the limited but highly informative data that can be used to correctly perform inference. 
From this point of view,  we need the agent to generate actions or trials that result in inferred user preference under which behavior of the environment is the same as its behavior under the true user preference $\bphi$. That is the same behavior over the history of taken actions and resulting observations. In what follows, we discuss and present our agent design.

\subsubsection{Agent Strategy}
Observe that behavior of the environment is characterized by a sequence of distributions, given by (\ref{eq:response_pred}), each indexed by the trials offered by the agent, since there is a  causal relation between trials and user responses. In order to generate an informative action, the agent has to first evaluate the environment, based on its knowledge from the previous interactions. Thus, the agent has to assign a predictive distribution $Q(r_n|r_1^{n-1},\{\bxr,\bxa\}_1^{n-1},\{\bxr_n,\bxa_n\})$ for the next response, in order to assess the informativeness of the trial $\{\bxr_n,\bxa_n\}$. We start with defining the measure that is used to assess the performance of the agent, and then present the agent strategy that results from optimizing this measure.

The self-information loss function, also called log-loss, given by $$-\log{Q(r_n|r_1^{n-1},\{\bxr,\bxa\}_1^{n-1},\{\bxr_n,\bxa_n\})},$$ characterizes the quality of the $Q$-distribution assignment. The problem of probability assignment that minimizes self-information loss  is  equivalent to the problem of universal prediction under self-information loss, see \cite{Merhav98universalprediction}.
In our case, though, prediction is not the main goal, but rather a mechanism to assess the validity of a statistical model and to further collect extra data (generate trial and observe the resulting user response) that can be used to improve the model.
Under the expected self-information loss criterion, i.e., $$\mathbb{E}[-\log{Q(R_n|r_1^{n-1},\{\bxr,\bxa\}_1^{n-1},\{\bxr_n,\bxa_n\})}],$$
where expectation is taken with respect to the generative user response distribution (we also call it true user or true source distribution), the optimal predictive assignment is actually the assignment of the true conditional probability $P(r_n|\bphi, r_1^{n-1}, \{\bxr,\bxa\}_1^{n-1},\{\bxr_n,\bxa_n\})$. Here the true source model that governs generation of $r_n$ belongs to the class of models, indexed by parameters $\bphi$,  given by $\{P(\cdot|\bphi,\cdot), \bphi\in \Omega\}$.  This true probability under certain ergodicity assumptions also minimizes the average self-information loss 
\begin{equation}
\frac{1}{n}\sum_{l=1}^n-\log{Q(r_l|r_1^{l-1},\{\bxr,\bxa\}_1^{n-1},\{\bxr_n,\bxa_n\})}.
\end{equation}
This sequential probability assignment gives rise to assignment of the probability $Q$ to the entire observation sequence $r_1^n$ and then sequential probability assignment and probability assignment problems for the whole sequence under self-information loss are equivalent. The average self-information loss for the true source distribution is, in fact, the entropy  $$H_n(P)=\mathbb{E}[-\log{P(R_1^n|\bphi,\{\bxr,\bxa\}_1^{n})}].$$
Observe that here we have conditioning on the trials, which are fixed agent's inputs. These trials $\{\bxr,\bxa\}_1^n$ can be modeled as side information, see e.g. \cite{CoverOrdentlich1996}, where side information is an additional information provided to the agent at the beginning of the next interaction. In our case, this side information is a causal function of previous interactions, see also (\ref{eq:agent_dist}).

In practice, the true source distribution is often unknown, and we have an extra loss, associated with using other than true distribution. This loss, which is coming on top of the minimum loss, is given by the information divergence between $P$ and $Q$, also called Kullback-Leibler (KL) divergence:
\begin{eqnarray}\nonumber
\mathbb{D}_n(P||Q)&\hspace{-5pt}\define&\hspace{-5pt}\mathbb{E}[-\log{Q(R_1^n|\{\bxr,\bxa\}_1^n)}-(-\log{P(R_1^n|\bphi,\{\bxr,\bxa\}_1^n)})]\\
&\hspace{-5pt}=&\hspace{-5pt}\mathbb{D}(P(R_1^n|\bphi,\{\bxr,\bxa\}_1^n)||Q(R_1^n|\{\bxr,\bxa\}_1^n)),
\end{eqnarray}
where expectation is taken with respect to $P.$
Clearly, to be able to produce informative trials, the agent needs sufficient  knowledge about the environment, and therefore should seek a probability assignment $Q$ that minimizes this KL-divergence.  

Recall that our agent has to perform universal prediction with respect to an indexed class of sources  $\{P(\cdot|\bphi,\cdot), \bphi\in \Omega\}$. However, since $\bphi$ is unknown, we take an approach, which has a strong Bayesian flavor, and assign some weighting distribution $w(\bphi)$ on $\bphi$. Then the quality of distribution assignment $Q$ will be assessed by the normalized weighted KL-divergence
\begin{eqnarray}\label{eq:DLcriterion}
\hspace{-12pt}D_n(Q,w,\{\bxr,\bxa\})&\define&\frac{1}{n}\int_{\Omega} w(\bphi)
\mathbb{D}_n(P||Q)d\bphi.
\end{eqnarray}

Next, we present the agent strategy together with its characterization and then show why this is an efficient strategy. 

\vspace{3pt}
\noindent {\textbf{Agent Strategy} \textit{ The agent strategy for the sequential trial design in the preference learning system, defined by distributions $\{P^a,P^u\}$, is to
\begin{enumerate}
\item apply the following distributions as generative agent and predictive user  distributions, respectively:
\begin{eqnarray*}\nonumber
&&\hspace{-20pt} P_l^a \,=\, p_l=\, \delta(\bxr_{l}-\bxa_{l-1})^{r_{l-1}}
\cdot \delta(\bxr_{l}-\bxr_{l-1})^{1-r_{l-1}}
\\\nonumber&&\hspace{30pt}
\cdot \hspace{-2pt}\int p(\bxa_l,\bgamma\,|\,r_1^{l-1},\{\bxr,\bxa\}_1^{l-1})\,d\bgamma\\
&&\hspace{-20pt} P_l^u = Q_{p,l}=\int_{\Omega} p(\bphi|r_1^{l-1},\{\bxr,\bxa\}_1^{l-1})
\\&&\hspace{40pt} 
\cdot P(r_l|\bphi,\{\bxr,\bxa\}_1^{l-1},\{\bxr_l,\bxa_l\})d\bphi,
\end{eqnarray*}
for $l=1,2,\ldots,n$, where
\begin{eqnarray*}
p(\bphi |  r_1^{l-1}, \{\bxr,\bxa\}_1^{l-1})
=\, \frac{p(\bphi)p(r_1^{l-1} |\bphi, \{\bxr,\bxa\}_1^{l-1})}{\int_{\Omega} p( \bphi') p(r_1^{l-1}|\bphi',\{\bxr,\bxa\}_1^{l-1})d\bphi'}
\end{eqnarray*}
is a posterior distribution and\\
$\int p(\bxa_l,\bgamma\,|\,r_1^{l-1},\{\bxr,\bxa\}_1^{l-1})\,d\bgamma$ 
is a marginal of the location parameter of this posterior;
\item generate the corresponding sequence of trials, for $l=2,3,$ 
$ \ldots, n$, based on the history of interactions with the user, according to the following rule
\begin{eqnarray}\nonumber
\{\bxr_l,\bxa_l\}&=&
\underset{
 \begin{matrix}
 \{\bxr,\bxa\}: \\
 P_l^a 
 \end{matrix}
 }{\arg \max}\hspace{-4pt}  I(R;\bphi\,|\,R_1^{l-1},\{\bxr,\bxa\}_1^{l-1}, \{\bxr,\bxa\}),
\end{eqnarray}
 where $I(\cdot;\cdot)$ is mutual information.
\end{enumerate}
This strategy is asymptotically optimal in that it minimizes normalized weighted KL-divergence. Given this strategy, for $n$ interactions, the achieved normalized weighted KL-divergence is given by
\begin{eqnarray}\nonumber
D_n(P^u,\{\bxr,\bxa\})&=&
\frac{1}{n} \sum_{l=1}^n I(R_l;\bphi|R_1^{l-1},\{\bxr,\bxa\}_1^{l-1}, \{\bxr_l,\bxa_l\}).
\end{eqnarray}
}}

\subsubsection{Predictive User Response Distribution}
In what follows, we discuss the principles guiding the agent strategy.
In order to assign a good predictive probability, the agent has to find  $w$ and $Q$ such that (\ref{eq:DLcriterion}) is minimized. One way to tackle the problem is to apply a maximin criterion that results in a mixture approach with a Bayesian flavor, i.e., 
\begin{eqnarray}
\sup_{w}\inf_{Q}D_n(Q,w,\{\bxr,\bxa\}).
\end{eqnarray}
For a given $w(\bphi),$  the minimizing distribution is given by the weighted distribution 
\begin{eqnarray}
Q_w(r_1^n|\{\bxr,\bxa\}_1^{n})=\int_{\Omega} w(\bphi)P(r_1^n|\bphi,\{\bxr,\bxa\}_1^n)d\bphi,
\end{eqnarray}
and the corresponding weighted divergence is given by $I_w(R_1^n;\bphi|\{\bxr,\bxa\}_1^{n}),$ where subscript $w$ indicates that some weighting distribution $w(\bphi)$ is used.
The agent, however, has to assign the predictive probabilities sequentially, after every interaction. Sequential assignment gives rise to the sequence probability assignment. It was shown in \cite{Matsushima1991} that  the above mixture of $$\{P(r_1^n|\bphi,\{\bxr,\bxa\}_1^{n}), \bphi\in \Omega\}$$ can be represented by the mixture of conditional probability functions
$$\{P(r_n|\bphi, r_1^{n-1},\{\bxr,\bxa\}_1^{n-1},\{\bxr_n,\bxa_n\}), \bphi\in \Omega\},$$
as 
\begin{eqnarray}\nonumber
&&\hspace{-25pt}Q_w(r_n|r_1^{n-1},\{\bxr,\bxa\}_1^{n-1},\{\bxr_n,\bxa_n\})\\\nonumber
&&=\int_{\Omega} w(\bphi|r_1^{n-1},\{\bxr,\bxa\}_1^{n-1})
\cdot P(r_n|\bphi,\{\bxr,\bxa\}_1^{n-1},\{\bxr_n,\bxa_n\})d\bphi,\\
\label{eq:univ_pu}
\end{eqnarray}
where the weighting function is a posterior distribution of $\bphi$ given $r_1^{n-1}$  and $\{\bxr,\bxa\}_1^{n-1}$, i.e.,
\begin{eqnarray}\nonumber
w(\bphi|r_1^{n-1},\{\bxr,\bxa\}_1^{n-1})
&=&p(\bphi |  r_1^{n-1}, \{\bxr,\bxa\}_1^{n-1})\\
&=&\frac{p(\bphi)p(r_1^{n-1} |\bphi, \{\bxr,\bxa\}_1^{n-1})}{\int_{\Omega} p( \bphi') p(r_1^{n-1}|\bphi',\{\bxr,\bxa\}_1^{n-1})d\bphi'},\label{eq:phi_posterior}
\end{eqnarray}
where the prior on the user preference parameters is independent of trials. Moreover, the performance of this sequential assignment is the same as the performance of the assignment when the whole sequence is observed. 
The minimizing distribution functions $Q_w$, in their turn, correspond to the normalized weighted information divergence given by 
\begin{eqnarray}\nonumber
D_n(Q_w,\{\bxr,\bxa\})
&=&\frac{1}{n}\sum_{l=1}^n\sum_{r_l\in \{0,1\}}\int_{\Omega} p(\bphi\,|\,r_1^{l-1},\{\bxr,\bxa\}_1^{l-1})\\\nonumber
&&\hspace{20pt}\cdot P(r_l\,|\,\bphi,r_1^{l-1},\{\bxr,\bxa\}_1^{l-1},\{\bxr_l,\bxa_l\})\\\nonumber
&&\hspace{20pt}\cdot \ln{\frac{P(r_l\,|\,\bphi,r_1^{l-1},\{\bxr,\bxa\}_1^{l-1},\{\bxr_l,\bxa_l\})}{P(r_l\,|\,\{\bxr,\bxa\}_1^{l-1}, \{\bxr_l,\bxa_l\})}}d\bphi\\
&&\hspace{-15pt}=\frac{1}{n}\sum_{l=1}^n I(R_l;\bphi|\,R_1^{l-1},\{\bxr,\bxa\}_1^{l-1}, \{\bxr_l,\bxa_l\})\label{eq:KLres}.
\end{eqnarray}
Here we have made a choice for the weighting functions up to the prior $p(\bphi)$, given by (\ref{eq:phi_posterior}). This is a sequential form of writing out $D_n(Q_w,\{\bxr,\bxa\})$ that coincides with the non-sequential result given by $I_w(R^n;\bphi|\{\bxr,\bxa\}_1^n).$

Note that in this universal prediction problem, the goal is to find a probability assignment that guarantees that the predictor performs well for all $\bphi$. The supremum of $1/nI_{w}(R^n;\bphi|\{\bxr,\bxa\}_1^n)$ over all allowable $w(\bphi)$ can be interpreted as a capacity  of the ``channel'' between the channel output $R_1^n$ and the channel input $\bphi$ with side information $\{\bxr,\bxa\}_1^n,$ defined by the class of sources $\{P(\cdot|\bphi,\cdot), \bphi\in \Omega\}.$ This capacity is given by 
$C_n \define 1/nI_{w^*}(R^n;\bphi|\{\bxr,\bxa\}_1^n),$ where $w^*(\cdot)$ is the capacity achieving prior, and then it holds that $\mathbb{D}_n(P||Q_{w^*})\leq n C_n$, $\forall \bphi$, guarantees in the minimax sense that
\begin{eqnarray}
\mathbb{D}_n(P||Q)&\geq& n C_n,  
\end{eqnarray}
On the other hand, the strong converse theorem of \cite{MerhavFeder1995} states that under self-information loss, for any asymptotically good approximation of $w^*(\phi)$, there is a remarkable concentration phenomenon, stating that, for any $\varepsilon>0$ and $w^*$-most values of $\bphi$, it holds that  
\begin{eqnarray}
\mathbb{D}_n(P||Q)&\geq& (1-\varepsilon)nC_n, \hspace{5pt} \forall Q. 
\end{eqnarray}
The practical problem is then finding the capacity achieving prior. However, the result in \cite{FederMehrav1996} allows $w(\cdot)$ to be an arbitrary weight function, thus telling  that $Q_w$ is optimal for $w$-most points in $\Omega$:
\begin{eqnarray}\label{eq:capacity_anyw}
\mathbb{D}_n(P||Q)&\geq& \mathbb{D}_n(P||Q_{w}).
\end{eqnarray}


Let's look at the interpretation of this result. This capacity gives information on the richness of the source class. The capacity-achieving weighting function assigns higher weights to the regions, where the sources are better discriminated and data is more informative about the parameters; and lower weights to the regions, where the sources are closer. Therefore, in the problem of preference learning, the use of posterior distribution of $\bphi$, as weighting functions, helps us to evaluate usefulness of the data generated by the agent for  inference of $\bphi$. We need to distinguish a single unknown source. Hence, on one hand, we aim at attaining the lower bound on the information divergence and, on the other hand, at finding good $w(\cdot)$ that further reduces this bound. In Section \ref{sec:UM_inference}, we further discuss approximation of the weighting distributions in their sequential form.

\subsubsection{Generative Agent Distribution}
Now that we have defined the predictor $Q_w$ to be used by the agent for preference learning system characterization, we need to specify the generative agent distribution. From  the  optimization point of view, the agent has to generate the proposals that lead to the optimal user preferences as quickly as possible. This means that, for every interaction with a user, the agent has to provide a better alternative than a reference proposal,
with high probability. This can be seen as a sort of gradient ascent method.  Clearly, this alternative would correspond to the user preference or the agent's estimate of it. Therefore, the generative distribution of the alternative proposal generation would coincide with the distribution of the optimal parameters corresponding to the the user preference. This is a marginal posterior distribution of location parameters.

Given the fact that the next reference proposal is just a deterministic function of the previous trial and user response to it,  the generative process for the next trial $l$ is given by
\begin{eqnarray}\nonumber
p( \{\bxr_l,\bxa_l\}| \bphi, r_1^{l-1}, \{\bxr,\bxa\}_1^{l-1})&&\\\nonumber  &&\hspace{-110pt}\define\,\delta(\bxr_{l}-\bxa_{l-1})^{r_{l-1}}\cdot \delta(\bxr_{l}-\bxr_{l-1})^{1-r_{l-1}}
\\
&&\hspace{-100pt} \cdot \int p(\bxa_l,\bgamma\,|\,r_1^{l-1},\{\bxr,\bxa\}_1^{l-1})\,d\bgamma,\label{eq:univ_pa}
\end{eqnarray}
where $p(\bxa_l,\bgamma\,|\,r_1^{l-1},\{\bxr,\bxa\}_1^{l-1})$ coincides with an expanded version of posterior (\ref{eq:phi_posterior}). Thus, the agent generative distribution functions $p_l^a$ are given by (\ref{eq:univ_pa}), for $l=2,3,\ldots,n.$

\subsubsection{Trial Design}
Next, we look at the problem of trial generation from a data informativeness point of view. The agent trial generation process is given by (\ref{eq:univ_pa}). The trials, generated by the agent, result in user responses, characterized by the predictor $Q_w$, which performance is given by the normalized weighted information divergence (\ref{eq:KLres}). As discussed above, this information divergence tells us how well data helps to distinguish the source, i.e., the individual user preference $\bphi.$ As such, we can also interpret this performance measure as the agent uncertainty about the user that we will call remaining system uncertainty (RSU). 

For the normalized weighted information divergence, trials serve as a causal side information and are given values produced by the agent. The generative agent distribution provides the ``direction'' of the data being sought for. However, to facilitate efficient learning of the optimal parameters, we also need to produce the most informative trials. To evaluate trial informativeness,  we use the normalized weighted information divergence, by looking at it as a function of trials. 

Since trials cause user responses, we gain the most information from getting a user response on the trial, about which our predictor is most uncertain. Thus, to generate the next trial at time $n,$ the agent should search for the worst case behavior of the optimal predictor from the trial (data) perspective, i.e.,

\begin{eqnarray}\nonumber
\{\bxr_n,\bxa_n\}&=&\hspace{-7pt}\underset{
\begin{matrix}
 \{\bxr_n,\bxa_n\}: \\
 P_n^a 
 \end{matrix}
 }{\arg\max} D_n(Q_w,\{\bxr,\bxa\})\\\nonumber
&&\hspace{-70pt}=\underset{
 \begin{matrix}
 \{\bxr_n,\bxa_n\}: \\
 P_n^a 
 \end{matrix}
 }{\arg\max}\,\, \sum_{l=1}^n I(R_l;\bphi|R_1^{l-1},\{\bxr,\bxa\}_1^{l-1}, \{\bxr_l,\bxa_l\})\\\nonumber 
&&\hspace{-70pt}=\underset{
 \begin{matrix}
 \{\bxr_n,\bxa_n\}: \\
 P_n^a 
 \end{matrix}
 }{\arg\max}\,\,  I(R_n;\bphi|R_1^{n-1},\{\bxr,\bxa\}_1^{n-1}, \{\bxr_n,\bxa_n\}),\\\label{eq:trial_design}
\end{eqnarray}
here due to sequential nature of interactions between the agent and the user,  the next trial to be generated by the agent corresponds to the argument of the maximum of the mutual information between the predicted response for a given next trial and optimal parameters, given the history of interactions. We call this last term under $\arg\max$  weighted individual divergence. This concludes the agent strategy for the trial design.

\section{Bayesian Inference for the User Model} \label{sec:UM_inference}
In the previous section, we defined the preference learning system and the experimental design strategy to obtain the data required for inference of the user model or, equivalently, parameters of the user preference function. The latter is, in fact, our main goal. Indeed,  user model inference and experimental design are intimately related. In  the experimental design, we are also concerned with user model inference, but there it is used as a weighting distribution. In general,  user and agent interactions are sequential, and, in the experimental design, the agent also assigns predictive probability sequentially.  Therefore, we concentrate and  present a sequential Bayesian procedure for user model inference.   

Recall that in the predictive probability assignment procedure, we have specified the weighting (posterior) distribution of the user model parameters up to the prior $p(\bphi)$. Therefore, we start here with this  prior specification and assume the following prior distribution for the user preference parameters:
\begin{equation}\label{eq:user_prior}
    p(\bphi)= \mathcal{N}\left(\boldsymbol{\phi}|\boldsymbol{\mu}_{\phi},\Sigma_{\phi}\right).
\end{equation} 
This assumption is the consequence of assumption that the user has one optimal preference, which means that preference distribution should have a clear optimum. 
Next, consider data points from the user-agent interactions in the preference learning system, denoted by $\mathcal{D}^{n-1}=\{r_l,\{\bxr_l,\bxa_l\}\}_{l=1}^{n-1}$. Given these data points, we work out  the posterior (\ref{eq:phi_posterior})  as
\begin{eqnarray}\nonumber
&&\hspace{-20pt}
p(\bphi \,|\,  r_1^{n-1}, \{\bxr,\bxa\}_1^{n-1})\\\nonumber 
&=& \frac{p(\bphi)p(r_1^{n-1}\,|\,\bphi,  \{\bxr,\bxa\}_1^{n-1})}{\int p( \bphi') p(r_1^{n-1}\,|\,\bphi', \{\bxr,\bxa\}_1^{n-1})d\bphi'}\\\nonumber
&=& p( \bphi)
\cdot \prod_{l=1}^{n-1}\frac{p(r_l \,|\, \bphi, r_1^{l-1}, \{\bxr,\bxa\}_l^{l-1},\{\bxr_l,\bxa_l\})}{P(D^l)}
\\\nonumber
&=& p( \bphi)
\cdot \prod_{l=1}^{n-1}\frac{p(r_l \,| \, \bphi, \{\bxr_l,\bxa_l\})}{P(D^l)}
\\\nonumber
&=& \mathcal{N}\left(\boldsymbol{\phi}|\boldsymbol{\mu}_{\phi},\Sigma_{\phi}\right)\cdot \prod_{l=1}^{n-1}\frac{1}{C_l}\cdot \Phi\left( f(\bxr_l;\bphi) - f(\bxa_l;\bphi) \right)^{r_l}\\
&& \cdot \Phi\left( f(\bxa_l;\bphi) - f(\bxr_l;\bphi) \right)^{1-r_l},
\end{eqnarray}
where $\mathcal{C}_{l}$ are  normalizing constants, and where, in the third equality, we used  the Markov property, since the user response only depends on the current trial, given parameters of the preference function. 
While this expression provides us with an analytical expression for the posterior, its exact estimation and use in the trial design is problematic, due to its non-linear form and complex integral evaluations. Therefore, to make the problem tractable, we turn to approximate sequential estimation, based on Assumed Density Filtering (ADF), also called moment matching, see, e.g., \cite{Opper1999} and \cite{Winther1999}.
The idea is to use a Gaussian distribution as an approximating distribution for the posterior. Note, that our assumption that a user has a single optimal preference means that the posterior should manifest it by a having a peak around true preference, which, in its turn, makes Gaussian approximation to be a viable assumption.  The ADF approach allows for a simple update of the first and second moments of the posterior after each trial. 

Following the ADF approach, we re-write the posterior distribution in the following way, by defining
$\tilde{p}_0(\boldsymbol{\phi})\define p(\boldsymbol{\phi}),$  $\tilde{p}_l(\boldsymbol{\phi})\define p(r_l|\boldsymbol{\phi},\{\bxr_l,\bxa_l\})$,  
\begin{eqnarray}
    p(\bphi|\mathcal{D}^{n-1}) 
    &=&  \prod_{l = 1}^{n-1}\frac{1}{\mathcal{C}_l}\cdot \tilde{p}_l(\boldsymbol{\phi}) \approx q_{n-1}(\boldsymbol{\phi}),
\end{eqnarray}
where $q_{n-1}(\boldsymbol{\phi})$ is our approximating family, which we choose to be a Gaussian one, i.e.,
\begin{eqnarray} 
q_{n-1}(\boldsymbol{\phi}) = \mathcal{N}\left(\boldsymbol{\phi} | \boldsymbol{\mu}_{\phi,{n-1}},\Sigma_{\phi,{n-1}}\right).
\end{eqnarray}
Observe that initially, at the beginning of interaction, we only have the prior parameter distribution and do not need any approximation. Then, after each trial,  we need to incorporate new evidence and perform an approximation. In order to do this, after each trial $l$, we take the ``exact'' posterior 
\begin{eqnarray}
    \hat{p}(\boldsymbol{\phi}|\mathcal{D}^l) 
    &=& \frac{\tilde{p}_l(\boldsymbol{\phi})q_{l-1}(\boldsymbol{\phi})}{\int{\tilde{p}_l(\boldsymbol{\phi})q_{l-1}(\boldsymbol{\phi})}d\boldsymbol{\phi}}
\end{eqnarray}
and find the approximating distribution $q_l(\cdot)$ that minimizes the KL-divergence:
\begin{eqnarray}
q_l(\bphi): \min_{q(\cdot)}\mathbb{D}(\hat{p}(\bphi|\mathcal{D}^l)||q(\bphi)),
\end{eqnarray}
given that $q_l(\cdot)$ belongs to the Gaussian approximation family. This problem is equivalent to the maximum likelihood estimation problem for $\hat{p}(\boldsymbol{\phi}|\mathcal{D}^l),$ and 
then the ADF procedure reduces to propagating the first two moments:
\begin{eqnarray}\label{eq:par_est_mu}
\boldsymbol{\mu}_{\phi,l}=\mathbb{E}_{q_l(\bphi)}\left[\bphi\right]&=&\mathbb{E}_{\hat{p}(\bphi|\mathcal{D}^l)}\left[\phi\right],\\
\nonumber
\Sigma_{\phi,l}&=&\mathbb{E}_{q_l(\bphi)}\left[(\bphi-\boldsymbol{\mu}_{\phi})^T(\bphi-\boldsymbol{\mu}_{\phi})\right]\\  &=&\mathbb{E}_{\hat{p}(\bphi|\mathcal{D}^l)}\left[(\bphi-\boldsymbol{\mu}_{\phi})^T(\bphi-\boldsymbol{\mu}_{\phi}) \right] .\label{eq:par_est_sig}
\end{eqnarray}
Thus, the problem of user model inference reduces to calculations of the moments of the ``true'' posteriors. Observe that the first moment, given by (\ref{eq:par_est_mu}-\ref{eq:par_est_sig}), provides us with the (current) estimate of the parameters of the user preference function. 

\section{Implementation of the Sequential Procedure for Active User Model Inference }
After specification of the experimental design and user model inference approach, we describe here implementation details for sequential active user model inference.

We start with initialization of the experimental design. To perform an initial trial, first two proposals to be evaluated are selected uniformly at random, i.e., $\mathbf{x}$ and $\mathbf{x}'$ according to $\mathrm{Uniform}\left([0,1]^D\right)$. Based on the user evaluation, they form the first trial $\{\bxr_1,\bxa_1\}.$
Furthermore, the model prior is given by (\ref{eq:user_prior}) as
\begin{equation*}
    p(\bphi)= \mathcal{N}\left(\boldsymbol{\phi}|\boldsymbol{\mu}_{\phi,0},\Sigma_{\phi,0}\right),
\end{equation*}
where $\boldsymbol{\mu}_{\phi,0},\Sigma_{\phi,0}$ are hyperparameters. 

After each interaction $l,$  $l=1,2,\ldots, n$, we obtain the user feedback on the trial, and thus the next data point $\{r_l, \{\bxr_l,\bxa_l\}\}$. At each step, we need to calculate the moments of the posterior distribution $\boldsymbol{\mu}_{\phi,l},\Sigma_{\phi,l}.$ To avoid calculations of integrals for non-linear distribution functions, we use Monte-Carlo simulations to infer the user model distribution. We apply the  Metropolis–Hastings (MH) sampling algorithm, for $\mathcal{D}_l$, according to the following model:
\begin{eqnarray}\label{eq:MH_par}
\bphi    & \sim & \mathcal{N}\left( \boldsymbol{\mu}_{\phi,l-1},\Sigma_{\phi,l-1}\right) \\\label{eq:MH_res}
r_l & \sim & \mathcal{B}{ernoulli} \left(\Phi( f(\bxa_l;\bphi) - f(\bxr_l;\bphi)\right).
\end{eqnarray}
The MH simulations result in a sequence of samples or posterior sample, denoted by $\tilde{q}_l(\boldsymbol{\phi})$. This sample is further used to calculate the corresponding moments, using maximum likelihood estimation:
\begin{eqnarray}\label{eq:mu_ml}
\boldsymbol{\mu}_{\phi,l}&=&\frac{1}{M}\sum_{\bphi_j\sim \tilde{q}_l(\bphi)}\! \! \bphi_j, \\\label{eq:sigma_ml}
\Sigma_{\phi,l}&=&\frac{1}{M-1}\sum_{\bphi_j\sim \tilde{q}_l(\bphi)}\! \! (\bphi_j-\boldsymbol{\mu}_{\phi,l})^T(\bphi_j-\boldsymbol{\mu}_{\phi,l}).
\end{eqnarray}
Hence, we obtain approximate posterior, given by $q_l(\boldsymbol{\phi})= \mathcal{N}\left(\bphi | \boldsymbol{\mu}_{\phi,l},\Sigma_{\phi,}\right)$, for $l=1,2,\ldots, n$.

To perform the trial design at step $l,$ we use the approximate posterior as a weighting distribution. Then, we need to calculate RSU in step $l$, given by the  mutual information in (\ref{eq:trial_design}). This mutual information expression can be further worked out as
\begin{eqnarray}\nonumber
I(R;\bphi\,|\,R_1^{l-1},\{\bxr,\bxa\}_1^{l-1}, \{\bxr,\bxa\})&\approx &\\
&&\hspace{-185pt}H(R\,|\,\mathcal{D}^{l-1}, \{\bxr,\bxa\})-\mathbb{E}_{q_l(\bphi)}\left[H\left(R\,|\,\bphi,\{\bxr,\bxa\}\right)\right],
\end{eqnarray}
where approximation is the result of using approximate posterior instead of the exact one.
Recall that our main goal is, in fact, inference of the location parameter in the user model. Therefore, we decided to treat the sensitivity parameter as a nuisance parameter. In this case, we have to modify the objective, given above, by integrating out the sensitivity  parameter and, hence, arrive at the following expression:
\begin{eqnarray}\nonumber
&&\hspace{-40pt}I(R;\halpha\,|\,R_1^{l-1},\{\bxr,\bxa\}_1^{l-1}, \{\bxr,\bxa\})\\\nonumber 
& \approx& \hspace{-4pt} H(R\,|\,\mathcal{D}^{l-1}, \{\bxr,\bxa\})\\
&&\hspace{-4pt}-\mathbb{E}_{q_l(\halpha)}\left[H\left(\mathbb{E}_{q_l(\bgamma\,|\,\halpha)}\left[R\,|\,\halpha,\bgamma,\{\bxr,\bxa\}\right]\right)\right]\hspace{-2pt}.\label{eq:MI_nuis}
\end{eqnarray}
Then, since our joint approximate posterior is Gaussian, which  first and second moments can be written as
\begin{eqnarray}
\boldsymbol{\mu}_{\phi,l} =
\begin{bmatrix} 
\boldsymbol{\mu}_{\alpha,l}\\ \boldsymbol{\mu}_{\gamma,l}
\end{bmatrix}, \hspace{5pt}
\Sigma_{\phi,l} =
\begin{bmatrix} 
\Sigma_{\alpha,l} & \Sigma_{\alpha\gamma,l} \\ 
\Sigma_{\gamma\alpha,l} & \Sigma_{\gamma,l}
\end{bmatrix},
\end{eqnarray}
the approximate marginal $q_l(\halpha)$ is also Gaussian
\begin{eqnarray}
q_l(\halpha)&=&\mathcal{N}(\halpha|\boldsymbol{\mu}_{\alpha,l},\Sigma_{\alpha,l}),
\end{eqnarray}
and so the conditional distributions  $q_l(\bgamma \,|\, \halpha),$  which expectation and covariance are given, according to the Markov-Gauss theorem, by
\begin{eqnarray}
\boldsymbol{\mu}_{\gamma\,|\, \alpha,l} &=&\boldsymbol{\mu}_{\bgamma,l} \,+\, \Sigma_{\gamma \alpha,l}\,  \Sigma_{\alpha,l}^{-1}\,\left(\halpha - \boldsymbol{\mu}_{\alpha,l}\right)\\
\Sigma_{\gamma| \alpha,l} &=& \Sigma_{\gamma,l} \,- \, \Sigma_{\gamma \alpha,l} \,\Sigma_{\alpha,l}^{-1}\, \Sigma_{ \alpha \gamma,l}.
\end{eqnarray}

Observe that decomposition of (\ref{eq:MI_nuis}) allows for easier computation of the mutual information, since the entropy terms are binary entropy functions, defined by $h(p)\define -p\log_2 p-(1-p)\log_2(1-p)$. Furthermore, we approximate expectations by averages and use approximate posterior to calculate the entropy terms in (\ref{eq:MI_nuis}) as follows
\begin{eqnarray}\nonumber    H(R\,|\,\mathcal{D}^{l-1},\{\bxr,\bxa\})
&=&h\left(p(r\,|\,\mathcal{D}^l,\{\bxr,\bxa\}\right)\\\nonumber 
&=& h\left( \int p(r\,|\,\bphi, \{\bxr,\bxa\}) \cdot p(\bphi\, |\,\mathcal{D}^{l-1}) \, d\bphi\right)\\ \nonumber 
&\approx& h\left(    
    \frac{1}{M}\cdot \hspace{-7pt}\sum_{\{\bphi_m\}_{m=1}^M \sim q_l(\bphi)} \hspace{-15pt}
       \Phi\left(\nu(\bphi_m)\right)\right)\\
    &=& h\left(    
    \frac{1}{M}\cdot \hspace{-10pt}\sum_{\{\bphi_m\}_{m=1}^M \sim \mathcal{N}\left(\bphi_m \,|\, \boldsymbol{\mu}_{\phi,l},\Sigma_{\phi,l}\right)} \hspace{-15pt}
       \Phi\left(\nu(\bphi_m)\right)\right), \label{eq:MI_t1}
    \end{eqnarray}   
where $\nu(\bphi_m)\define f(\bxr;\bphi_m) - f(\bxa;\bphi_m);$   
\begin{eqnarray}\nonumber 
&&\hspace{-20pt} \mathbb{E}_{q_l(\halpha)}\left[H\left(\mathbb{E}_{q_l(\bgamma\,|\,\halpha)}\left[R\,|\,\halpha,\bgamma,\{\bxr,\bxa\}\right]\right)\right]    
\\\nonumber 
&\approx& \frac{1}{M} \sum_{\{\halpha_m\}_{m=1}^M \sim q_l(\halpha)}  
h\left(\frac{1}{M'}\sum_{\{\bgamma_{m'}\}_{m'=1}^{M'}\sim q_l(\bgamma\,|\,\halpha_m)}\hspace{-30pt}\Phi\left(  \nu(\halpha_m,\bgamma_{m'})\right)\right)\\\nonumber
&=&\frac{1}{M}\cdot\\\nonumber
&&\hspace{-10pt}\sum_{\{\halpha_m\}_{m=1}^M \sim \mathcal{N}\left(\halpha_m \,|\, \boldsymbol{\mu}_{\alpha,l},\Sigma_{\alpha,l}\right)} 
\hspace{-10pt}h\left(\frac{1}{M'}\sum_{\{\bgamma_{m'}\}_{m'=1}^{M'}\sim \mathcal{N}\left(\bgamma_{m'} \,|\, \boldsymbol{\mu}_{\gamma\, | \,\alpha,n},\Sigma_{\gamma\, | \,\alpha,l}\right)}\hspace{-30pt}\Phi\left(  \nu(\halpha_m,\bgamma_{m'})\right)\right),\\\label{eq:MI_t2}
\end{eqnarray}
where $\nu(\halpha_m,\bgamma_{m'})$ is defined in a similar way as above, but with expanded parameters, see also (\ref{eq:parame_trans}).

Next, the generative agent distribution for the trials is given by
\begin{eqnarray}
p_l^a &=&\delta(\bxr_{l}-\bxa_{l-1})^{r_{l-1}}\cdot \delta(\bxr_{l}-\bxr_{l-1})^{1-r_{l-1}}
\cdot \mathcal{N}(\bxa_l\,|\,\boldsymbol{\mu}_{\alpha,l},\Sigma_{\alpha,l}).
\end{eqnarray}
Therefore, generating trials from this distribution is equivalent to setting 
\begin{eqnarray}
\bxr_{l}&=& (\bxa_{l-1})^{r_{l-1}} \cdot (\bxr_{l-1})^{1 - r_{l-1}}
\end{eqnarray}
and generating an alternative proposal according to
\begin{eqnarray}\label{eq:proposal}
\bxa_l &\sim& \mathcal{N}(\bxa_l\,|\,\boldsymbol{\mu}_{\alpha,l},\Sigma_{\alpha,l}).
\end{eqnarray}

Finally, given the expressions above, the trial design constitutes generating a pool of alternative proposals, using  (\ref{eq:proposal}),  calculating mutual information terms for the resulting trials, using (\ref{eq:MI_t1}) and (\ref{eq:MI_t2}), and selecting the alternative that corresponds to the highest mutual information term. This procedure is also given in Alg.~\ref{alg:trial}, where $\MI(\cdot)$ is computed, using (\ref{eq:MI_t1}) and (\ref{eq:MI_t2}).

\begin{algorithm}[!tbh]
\caption{Trial design}
\label{alg:trial}
\KwData{
    \begin{itemize}
        \item $q_{l-1}(\bphi)$ -- approximate posterior after $l-1$ interactions
        \item $\{\bxr_{l-1}, \bxa_{l-1}\}$ -- trial at $l-1$-th interaction
        \item $r_{l-1}$ -- reply to $l-1$-th trial
        \item $M,M'$ -- number of Monte-Carlo samples
        \item $N$ -- number of samples for alternative proposal generation
    \end{itemize}
}
\KwResult{ 
\begin{itemize}
    \item $\{\bxr_{l}, \bxa_{l}\}$ -- new trial
    \item $D_l$ --  system uncertainty or weighted individual KL-divergence of $l$-th step
\end{itemize}
}
\BlankLine

$\bxr_{l} \leftarrow (\bxa_{l-1})^{r_{l-1}} \cdot (\bxr_{l-1})^{1 - r_{l-1}}$\;
$\bxa_l \leftarrow $\Sample{$q_{l-1}(\bphi)$}\;
$D_l \leftarrow$\MI{$M,M', q_{l-1}(\bphi), \bxa_l, \bxr_{l}$}\;

\For{$i \leftarrow 1$ \KwTo $N-1$}{
    $\bxa \leftarrow$\Sample{$q_{l-1}(\bphi)$}\;
    $D \leftarrow$\MI{$M,M', q_{l-1}(\bphi), \bxa, \bxr_{l}$}\;
    \If{$D_l < D$}{
        $\bxa_l \leftarrow \bxa$\;
        $D_l \leftarrow D$
    }
}
\end{algorithm}

Our learning procedure can be summarized as a sequential process of trials and user model inferences, with the stopping criterion, determined by convergence of the normalized weighted  KL-divergence, in a sense that convergence is achieved, when uncertainty reduction becomes smaller than some small predefined $\delta.$ The learning procedure is also provided as a pseudo-code in Alg.~\ref{alg:learn}. There \MC{$\cdot$} denotes Monte-Carlo simulations according to the model, given by  (\ref{eq:MH_par})-(\ref{eq:MH_res}),  
followed by calculations in (\ref{eq:mu_ml}) and (\ref{eq:sigma_ml}).

\begin{algorithm}[!htb]
\caption{Learning procedure}
\label{alg:learn}
\KwData{
    \begin{itemize}
        \item $q_0(\bphi)$ -- model prior
        \item $\{\bxr_{1}, \bxa_{1}\}$ -- initial trial 
        \item $r_{1}$ -- reply to the initial trial
        \item $M,M'$ -- number of Monte-Carlo samples
    \end{itemize}
}
\KwResult{
    \begin{itemize}
    \item $q(\bphi)$ -- user model parameter distribution
    \item $\mathbb{E}_{q}(\halpha)$ -- optimal parameter estimate
    \item $D$ -- remaining system uncertainty 
    \end{itemize}
}
\BlankLine

$l \leftarrow 0$\;
$D_l\leftarrow \infty$\;
$D\leftarrow 0$\;
\While {$D_l> l\delta$}{
$l \leftarrow l+1$\;
$\{\bxr_{l},\bxa_{l}\}, D_l \leftarrow$ \Design{$q_{l-1}(\bphi), \{\bxr_{l-1},\bxa_{l-1}\},r_{l-1}$}\;
$D \leftarrow D+D_l$ \;
$r_l \leftarrow$ \Trial{$\{\bxr_{l},\bxa_{l}\}$}\;
$q_l(\bphi) \leftarrow$ \MC{$M, \boldsymbol{\mu}_{\phi,l-1},\Sigma_{\phi,l-1}, \{\bxr_{l},\bxa_{l}\}, r_l$}
}
$q(\bphi) \leftarrow q_{l}(\bphi)$\;
$\mathbb{E}_{q}(\halpha)\leftarrow \boldsymbol{\mu}_{\alpha,l}$ \;
$D \leftarrow D/l$ \;
\end{algorithm}
\clearpage 

\section{Experimental Results}
\subsection{Simulation Results}
In this section, we present simulation results for a toy example of user preference learning. This example allows us to visualize important aspects of the agent performance.  The user model is given by Def.~\ref{def:usermodel}, where two synthetic parameters have to be learned.   We also simulate user responses by sampling them according to the user model, in order to create the whole interaction cycle of the user preference system, see Fig.~\ref{fig:ha_sys}. In our simulations, we know the true values of the user preference function, and our goal is to investigate a number of aspects of the proposed approach. First, we are interested to show feasibility of learning the true preferences by showing that estimated user preferences for synthetic parameters converge to the true user preferences for a simulated user. Moreover, since, in practice, user preferences are unknown, our goal is to validate that RSU  provides a good metric for performance monitoring of  user preference learning. As a benchmark, we use root-mean-squared error (RMSE), that can be used when the true preference is known. 

The numerical experiments were implemented in Python, where probabilistic modeling was realized in  probabilistic
programming language (PPL), using the Theano-PyMC 1.1.2. In particular, it was used to implement the Hamiltonian Monte-Carlo sampling in Alg.~\ref{alg:learn}. We used a desktop computer with Intel i7 CPU Core and 32 GB of RAM. 
Under these conditions, one full interaction step took about 12sec.

\begin{figure*}[!htb]
   \centering
   \begin{subfigure}{0.9\textwidth}
   \includegraphics[scale=0.7]{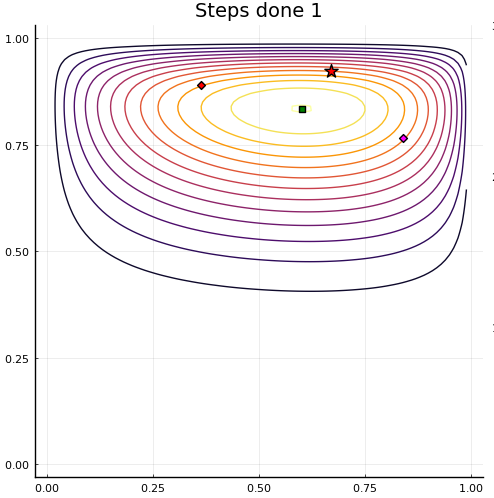}
   \caption{}
   \end{subfigure}
   \begin{subfigure}{0.9\textwidth}
   \includegraphics[scale=0.7]{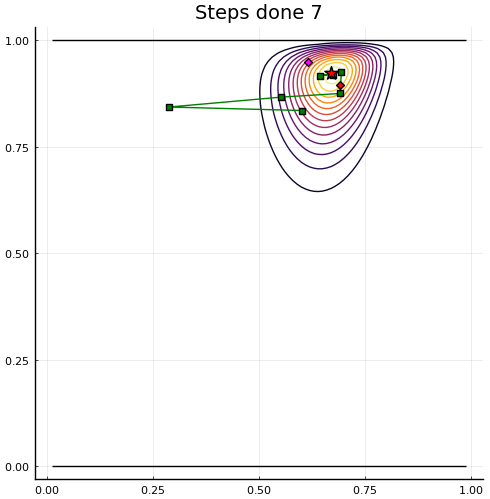}
   \caption{}
   \end{subfigure}
   \end{figure*}

   \begin{figure*}[!thb]\ContinuedFloat
   \begin{subfigure}{0.9\textwidth}
   \includegraphics[scale=0.7]{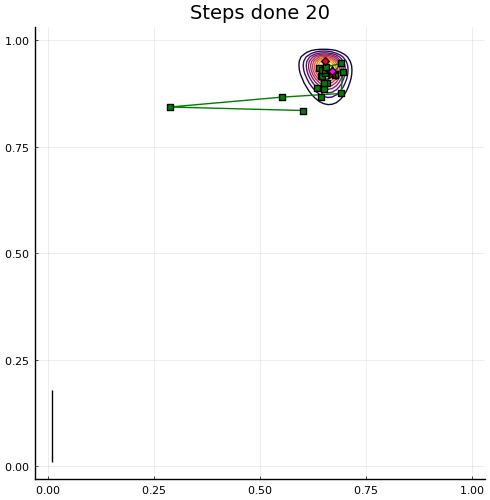}
   \caption{}
   \end{subfigure}
    
   \begin{subfigure}{0.9\textwidth}
    \includegraphics[scale=0.7]{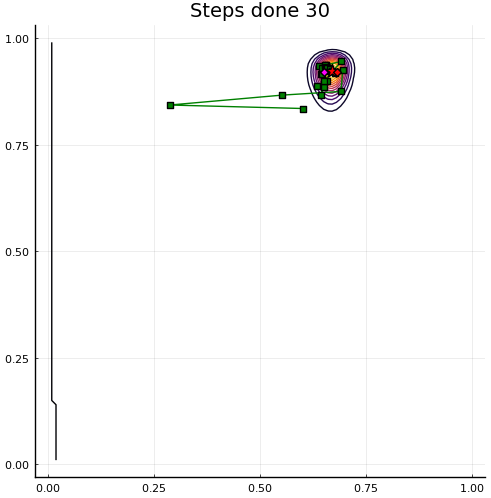}
    \caption{}
   \end{subfigure}
   \end{figure*}

   \begin{figure*}[!thb]\ContinuedFloat
   \begin{subfigure}{0.9\textwidth}
    \includegraphics[scale=0.7]{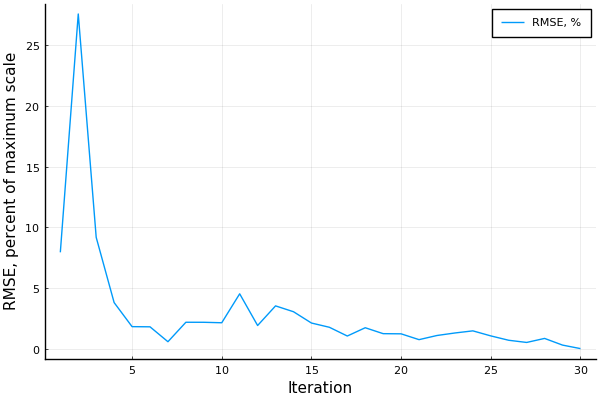}
    \caption{}
     \end{subfigure}
    \begin{subfigure}{0.9\textwidth}
    \includegraphics[scale=0.7]{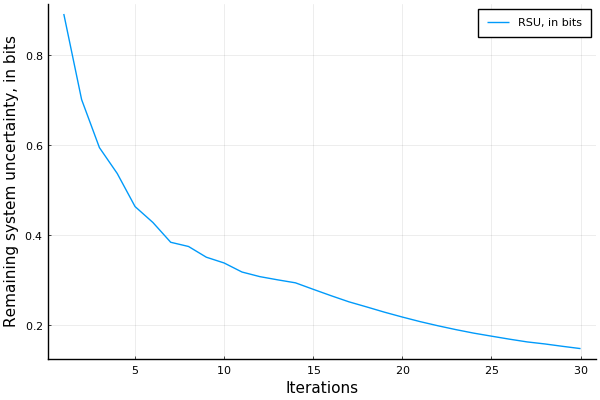}
    \caption{}
   \end{subfigure}
   \vspace{-10pt}
   \caption{The inferred synthetic parameter distribution after 1, 7, 20 and 30 interactions in (a)-(d) and the corresponding RMSE, in (e), and RSU, in (f). The red star denotes the true preference, green dots and connecting lines denote the trajectory of the preference estimates before each interaction, the red dot corresponds to the current user preference from the last trial and the magenta dot to the alternative proposal. }
   \label{fig:dist_beh}
   \end{figure*}

\begin{figure*}[!htb]
   \centering
   \begin{subfigure}{0.9\textwidth}
   \includegraphics[scale=0.7]{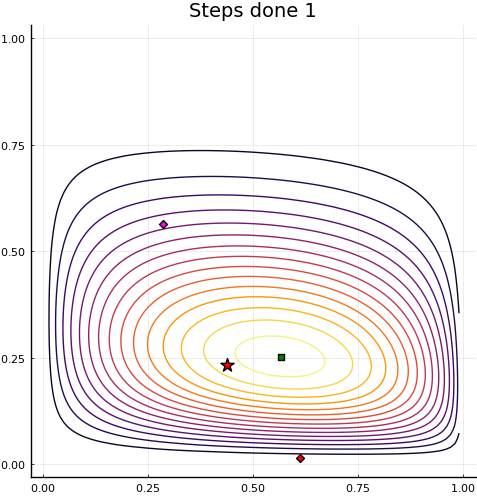}
   \caption{}
   \end{subfigure}
   \begin{subfigure}{0.9\textwidth}
    \includegraphics[scale=0.7]{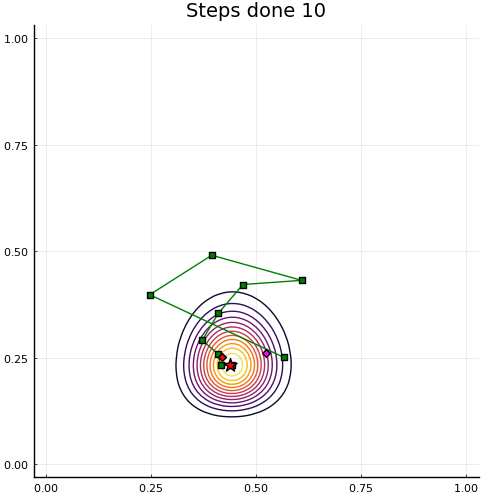}
    \caption{}
   \end{subfigure}
   \end{figure*}

   \begin{figure*}[!thb]\ContinuedFloat
   \begin{subfigure}{0.9\textwidth}
    \includegraphics[scale=0.7]{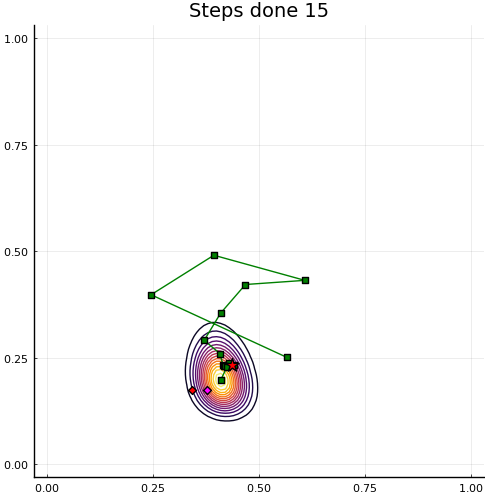}
    \caption{}
   \end{subfigure}
   \begin{subfigure}{0.9\textwidth}
    \includegraphics[scale=0.7]{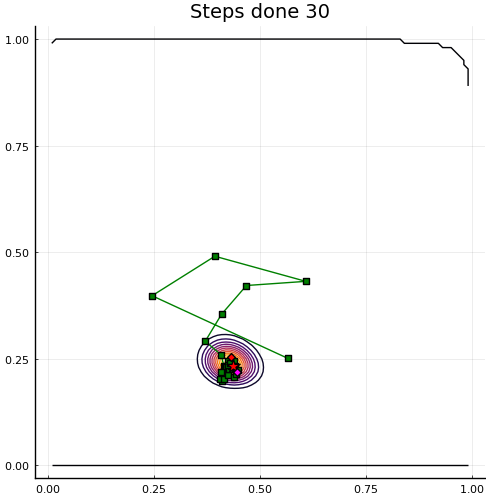}
    \caption{}
   \end{subfigure}
   \end{figure*}

   \begin{figure*}[!thb]\ContinuedFloat
   \begin{subfigure}{0.9\textwidth}
    \includegraphics[scale=0.7]{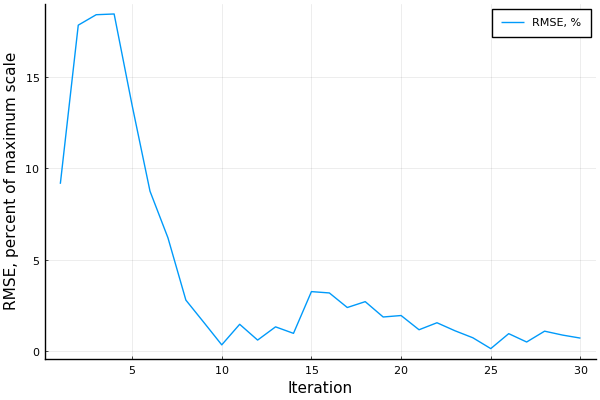}
    \caption{}
     \end{subfigure}
    \begin{subfigure}{0.9\textwidth}
    \includegraphics[scale=0.7]{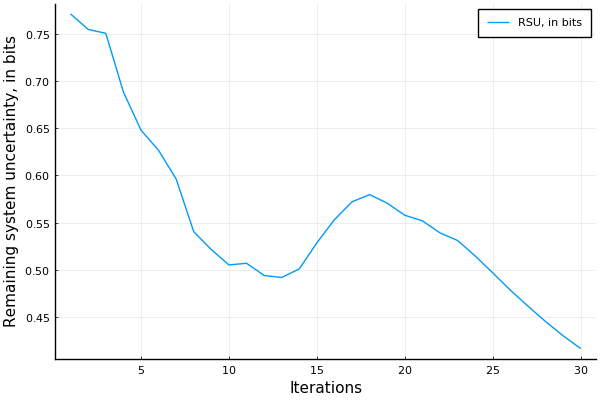}
    \caption{}
   \end{subfigure}
    \caption{The inferred synthetic parameter distribution after one, 10, 15 and 30 interactions,  in (a)-(d), and the corresponding RMSE, in (e), and RSU, in (f). Notations in (a)-(d) are the same as in Fig.~\ref{fig:dist_beh}}
    \label{fig:dist_beh_fl}
\end{figure*}
 
\begin{figure*}[!htb]
\centering
\begin{subfigure}{0.9\textwidth}
   \includegraphics[scale=0.7]{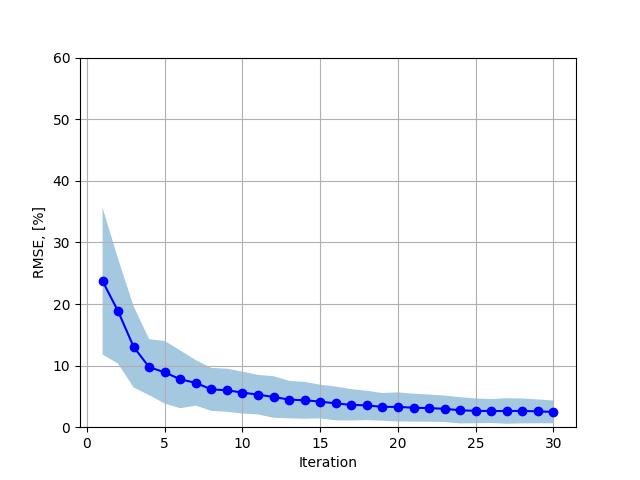}
    \caption{Root-mean-squared error (RMSE)}
    \label{fig:average_error}
\end{subfigure}
\begin{subfigure}{0.9\textwidth}
   \includegraphics[scale=0.7]{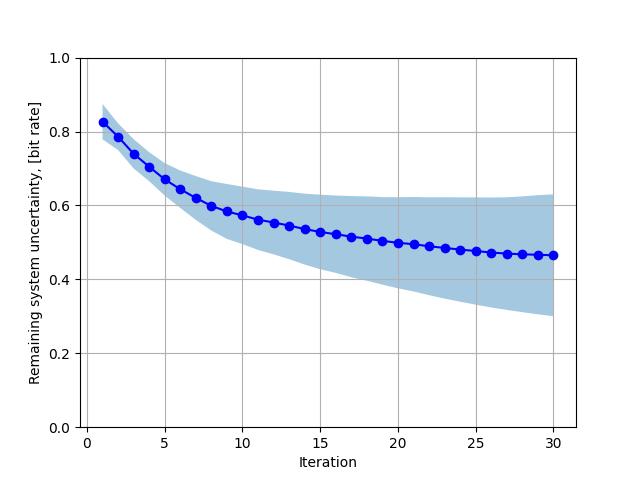}
    \caption{Remaining system uncertainty (RSU)}
    \label{fig:rsu}
\end{subfigure}
\caption{RMSE and RSU with $1\sigma$ confidence intervals}
\end{figure*}
\clearpage
Fig. \ref{fig:dist_beh} shows the true optimal parameters, their estimates and the proposals, generated by our agent together with the estimated parameter distribution. Note that as more and more data becomes available, the posterior distribution converges and become very dense around the optimum, and thus the proposals would eventually coincide with or become very close to the true optimum. At the early stage of learning, though, the proposals need not be close to the optimum. At this stage, the exploration process is dominating, where the agent generates proposals that aim at largest reduction of the uncertainty about a user. Indeed, looking at Fig. \ref{fig:dist_beh}, we see that the alternative proposals at the beginning of the learning process are spread further away from the estimated distribution mean than at the later stages of learning.  In general, we see that the learning procedure of the optimum user preferences converges quickly to the true optimum. Moreover, if we look at RMSE and the agent uncertainty about the user,  we see that both RMSE and RSU decrease with time, show-casting quick learning performance of our approach. We run the simulations for 30 interactions, but we see that a good estimate can be achieved with less than 20 interactions.  While RSU follows the trend of RMSE, FRMSE is more sensitive to inconsistencies in the point estimator. RSU is smoother, since it is a functional of the user response distribution. Moreover, as a functional, it provides more information about the learning behavior than a point estimator. Observe that, in Fig. \ref{fig:dist_beh_fl}, RSU goes up in situations when the user becomes less consistent in his responses, but in the end stabilizes as the responses become more reliable and, consequently, the distribution estimate becomes better. This demonstrates that RSU can efficiently be used to monitor the preference learning performance in the field.

Statistical behavior of the performance metrics, RMSE and RSU, is shown in
Fig.~\ref{fig:average_error} and Fig.~\ref{fig:rsu}, respectively.  These results were obtained by  simulating the learning procedure $T=100$ times, where each time a new true user preference was selected at random, and simulating $n=30$ interactions in each learning cycle. Note that due to stochastic exploration nature of the agent strategy, RSU variance increases with the interactions. Indeed, in the beginning of learning reduction of uncertainty is the largest, resulting into small RSU variance. At the later stage, exploration paths can vary with occasional increase in RSU, resulting thus into larger variability. 

Figure \ref{fig:other_appraoches} presents the results of applications of Gaussian processes with different acquisition functions, i.e., expected improvement, upper confidence bound (UCB) and Thomson sampling. Looking at these results, we see that our approach outperforms the resulting methods. After 30 interactions our agent strategy achieves less then 3\% RMSE, while the GP-based methods can only achieve from 8 to 11\% RMSE on average with rather large variance.

\begin{figure*}[!htb]
\centering
\begin{subfigure}{0.9\textwidth}
   \includegraphics[scale=0.7]{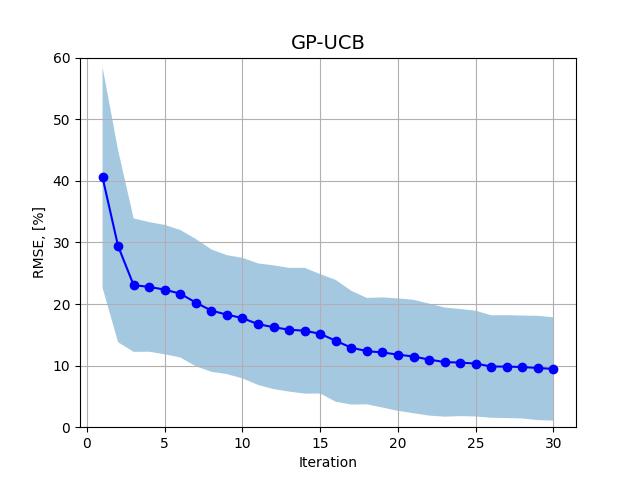}  
\end{subfigure}
\end{figure*}
\begin{figure*}[!thb]\ContinuedFloat
\begin{subfigure}{0.9\textwidth}
   \includegraphics[scale=0.7]{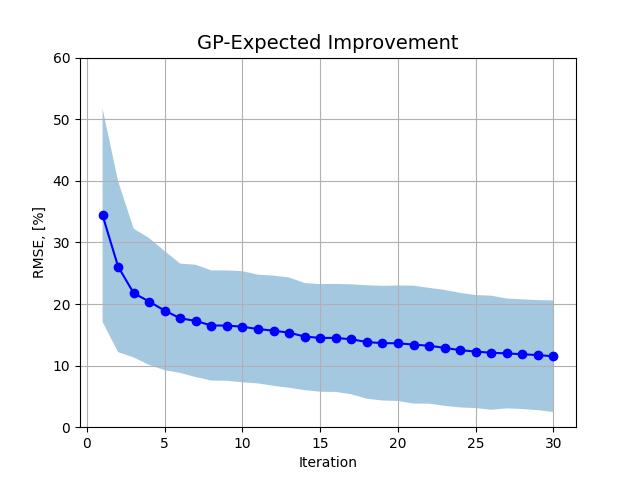}
\end{subfigure} 
\begin{subfigure}{0.9\textwidth}
   \includegraphics[scale=0.7]{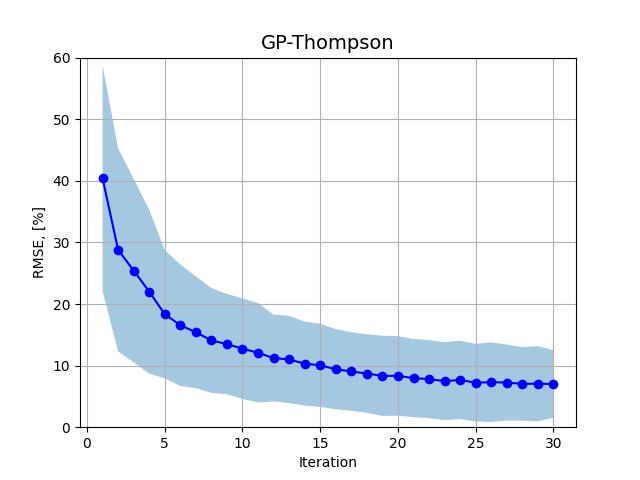}
\end{subfigure}
\caption{RMSE with $1\sigma$ confidence intervals for GP-based agents}
\end{figure*}\label{fig:other_appraoches}
\clearpage


\subsection{HA Application Example}
In this section discuss the results of applying our proposed preference learning approach to the problem of fitting and tuning HAs. We consider the situation of tuning gains and noise suppression level in HAs, resulting in 11 HA parameters that we would like to optimize by conducting the trials with hearing impaired users.

For the user-agent interactions, the users were presented with two sound samples processed with two sets of parameters. The users have been instructed to think about both their comfort and ability to understand the speech, while evaluating the proposals. The interactions continued until the users were satisfied with the settings and could not perceive the differences in the presented sound samples. We have conducted three study session per user, where three different initial sound settings were used: professionally selected  and tuned settings (we call it ``experienced''), settings based on self-measured hearing loss (we call it ``first-fit''), and random settings (called simply ``random'').

In order to evaluate the performance of the preference learning, speech recognition task was performed for initial and learned conditions. During this task, the user were presented with 50 non-sense sentences (different for initial and learned conditions) from the BEL sentence corpus \cite{BEL-data,Bel-development}. The task was intentionally difficult and did not allow to infer possibly missing words from the context. 

Fig. \ref{fig:RSU_HA} shows RSU and number of steps before we consider the  user to find his/her preferences. Given 11 parameters that we optimize, the results look very promising as it only requires the users maximum 24 steps to obtain satisfying HA settings. Note that the number of steps does not have a dependency on the initial conditions. For Subject 1, e.g., it took 24 steps to fine-tune the parameters, when the subject started with good, professionally selected settings; while Subject 2only had to do 11 interactions for the same conditions.    

\begin{figure*}[!htb]
\centering
\begin{subfigure}{0.9\textwidth}
   \includegraphics[scale=0.7]{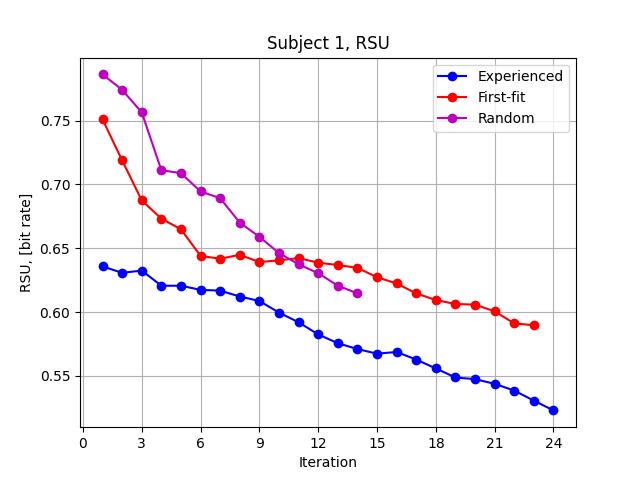}  
\end{subfigure}
\begin{subfigure}{0.9\textwidth}
   \includegraphics[scale=0.7]{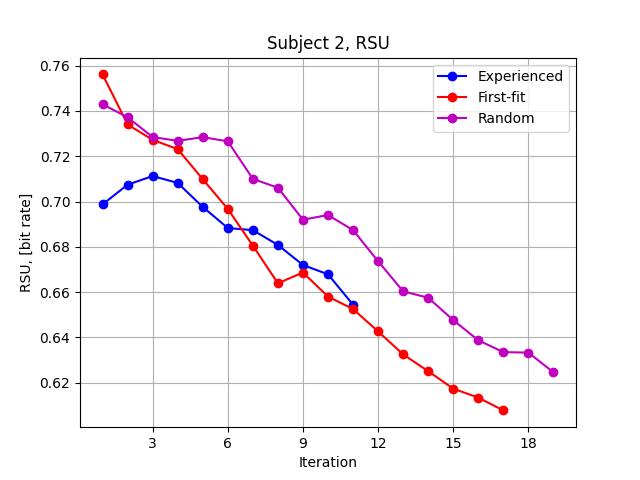}
\end{subfigure} 
\end{figure*}
\begin{figure*}[!thb]\ContinuedFloat
\begin{subfigure}{0.9\textwidth}
   \includegraphics[scale=0.7]{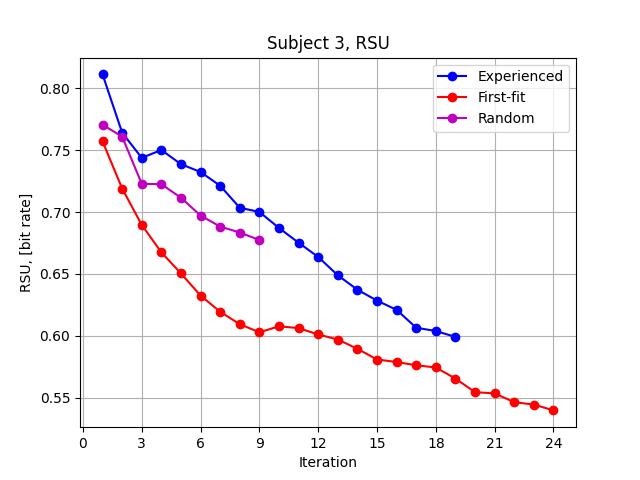}
\end{subfigure}
\begin{subfigure}{0.9\textwidth}
   \includegraphics[scale=0.7]{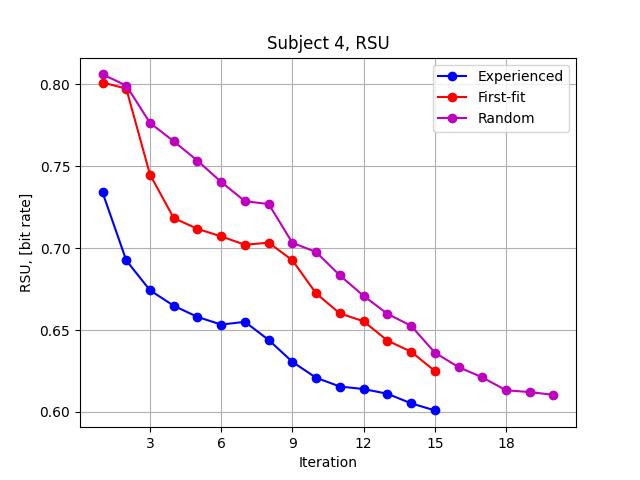}
\end{subfigure}
\caption{RSU in HA settings preference learning study}
\end{figure*}\label{fig:RSU_HA}
\clearpage

Next, we look at the speech recognition performance. While leaned HA settings are perceived as pleasant and desirable, we are also interested in speech recognition performance delivered by the learned settings. Table \ref{tab:speech_imp} presents speech improvement results for different initial conditions. The results are average improvement over 50 sentences, measured as the number of correctly recognized keywards in the sentences; word error rate (WER); and word information preserved (WIP). The latter two metrics are reported, since speech recognition tools were used to transcript the sentences repeated by the users. We observe improved speech recognition for all initial conditions. Clearly, improvement is smallest when initial conditions were already quite good, i.e., selected and fine-tuned by a professional. Nevertheless, even for this condition we achieve both speech recognition improvement and higher sound quality satisfaction by the users. 

\begin{table}[]
\begin{center}			
\begin{tabular}{l|ccc}
	 &Keywords &	 WER &	 WIP\\
  \hline
Experienced &	5.21 &	5.55&	2.18\\
Self-fit &	28.07&	19.76&	38.33\\
Random	&24.95&	13.69&	23.48\\
\end{tabular}	
\end{center}\caption{Measured speech improvement, in \%}\label{tab:speech_imp}	
\end{table}

Finally, using the results presented in  Table \ref{tab:speech_perf}, we analyze the difference in the speech recognition performance, measured as average number of correct keywords (Keywords), and WER and WIP, when different initial conditions were used. We see that the resulting speech recognition performance is comparable for all the conditions and are all better than fine-tuned professional settings. Thus, looking at all these results, we can conclude that our preference learning approach is a promising technique to be applied in the hearing aid industry. 

\begin{table}[]
\begin{center}			
\begin{tabular}{l|ccc}
	 &Keywords &	 WER &	 WIP\\
  \hline
Experienced initial&	2.1207&	0.5040 &	0.3839\\
Experienced learned &	2.2481&	0.4816&	0.3934\\
Self-fit learned&	2.5489&	0.4085&	0.4687\\
Random	learned&2.2713&	0.4775&	0.3969\\
\end{tabular}	
\end{center}\caption{Speech recognition performance}\label{tab:speech_perf}	
\end{table}

\section{Discussions and Conclusions}
In this paper, we have studied the problem of active preference learning using pairwise comparison. We have modeled preference learning system as two interacting sub-systems, one representing a user with his preferences for some outcome, characterized by a set of parameters, and another one representing the agent that guides the learning process by offering informative trials. In order to monitor the learning behavior and realize the agent strategy for obtaining  efficient data from interactions with a user, we have proposed to use the weighted normalized information divergence. This divergence characterizes the agent uncertainty about the user. Optimization of this performance measure results into an optimal strategy for the data collection and guarantees efficient accurate learning of the user preference.  Simulation results show that the remaining system uncertainty can be used to monitor the learning behavior of the agent in practice, when the true user preference is unknown. Moreover, we also see that the agent trial generation, based on this metric, results in quick reduction of the search space for the optimal user parameter preferences and their accurate estimation. Also user study on application of our approach to HA personalization demonstrated quick convergence to good HA settings that deliver speech improvement and better sound quality.

Our work has a close connection to universal prediction, as well as universal source coding. Here, however, our goal, when assigning a predictive distribution, is user model quality evaluation. Moreover, the problem we study relates to the active inference problem, see e.g. \cite{Friston2009}.  However, we take somewhat different approach at learning.  Instead of putting a focus on designing trials that minimize uncertainty about (future) user responses and thus aiming at supporting the current estimates, we look at the trials that result in the largest uncertainty in the user response and solicit responses to correct the estimate and resolve the uncertainty. Approximate  inference, i.e.,  ADF, plays a role in our work, when we infer the parameters of the user preference function, given actively generated and observed data. Observe that sequential or online nature of the proposed learning approach allows continuous learning of dynamic user preferences in situations when they change over time. Finally, we would like to mention that preference learning has a similar in flavor to reinforcement learning: both are processing the feedback from the environment and have to search for the optimal action strategy. Nevertheless, these two disciplines are conceptually different. First, in reinforcement learning, the goal is known and the reward is formalized using this goal. This is not the case in the preference learning.  Next, the state of environment - preference in our case, is unknown and unobservable in preference learning, while reinforcement learning relies on the state observability.

\section*{Acknowledgements}
We would like to thank Yuri Kirnos for his invaluable help in setting up the user testing system for HA user study; Tao Cui for his help in administrating the study; Greg Olsen and Michalis Papakostas for all their help and support of the user study. 

\bibliography{zotero}

\end{document}